\documentclass[conference]{IEEETran}
\usepackage[utf8]{inputenc} 
\usepackage[T1]{fontenc}    
\usepackage{hyperref}       
\usepackage{url}            
\usepackage{booktabs}       
\usepackage{amsfonts}       
\usepackage{nicefrac}       
\usepackage{microtype}      
\usepackage{xcolor}         
\usepackage{amsmath,amssymb}
\usepackage{wrapfig}
\usepackage{graphicx}
\usepackage{float}
\usepackage{caption}
\usepackage{subcaption}
\usepackage{bm}
\usepackage{listings}

\newcommand{\Rb}{\mathbb{R}}

\NewDocumentCommand{\codeword}{v}{%
\texttt{\textcolor{blue}{#1}}%
}

\title{TKAN: Temporal Kolmogorov-Arnold Networks}
\author{%
  \IEEEauthorblockN{%
    Rémi Genet\IEEEauthorrefmark{1}\textsuperscript{\textsection} and
    Hugo Inzirillo\IEEEauthorrefmark{2}\textsuperscript{\textsection}
  }%
  \IEEEauthorblockA{\IEEEauthorrefmark{1} DRM, Université Paris Dauphine - PSL}%
  \IEEEauthorblockA{\IEEEauthorrefmark{2} CREST-ENSAE, Institut Polytechnique de Paris}%
}

\begin{document}
\thispagestyle{plain}
\pagestyle{plain}
\maketitle

\begingroup\renewcommand\thefootnote{\textsection}
\footnotetext{These authors contributed equally.}
\endgroup

\begin{abstract}
Recurrent Neural Networks (RNNs) have revolutionized many areas of machine learning, particularly in natural language and data sequence processing. Long Short-Term Memory (LSTM) networks have demonstrated their ability to capture long-term dependencies in sequential data. Inspired by the Kolmogorov-Arnold Networks (KANs), a promising alternative to Multi-Layer Perceptrons (MLPs), we propose a new neural network architecture inspired by KAN and LSTM, called ``Temporal Kolomogorov-Arnold Networks'' (TKANs). TKANs combine the strength of both networks. They are composed of Recurring Kolmogorov-Arnold Networks  (RKANs) Layers embedding memory management. This innovation enables us to perform multi-step time series forecasting with enhanced accuracy and efficiency. By addressing the limitations of traditional models in handling complex sequential patterns, the TKAN architecture offers significant potential for advancements in fields requiring more than one step ahead forecasting.

\begin{figure}[H]
    \centering
    \includegraphics[width=0.8\linewidth]{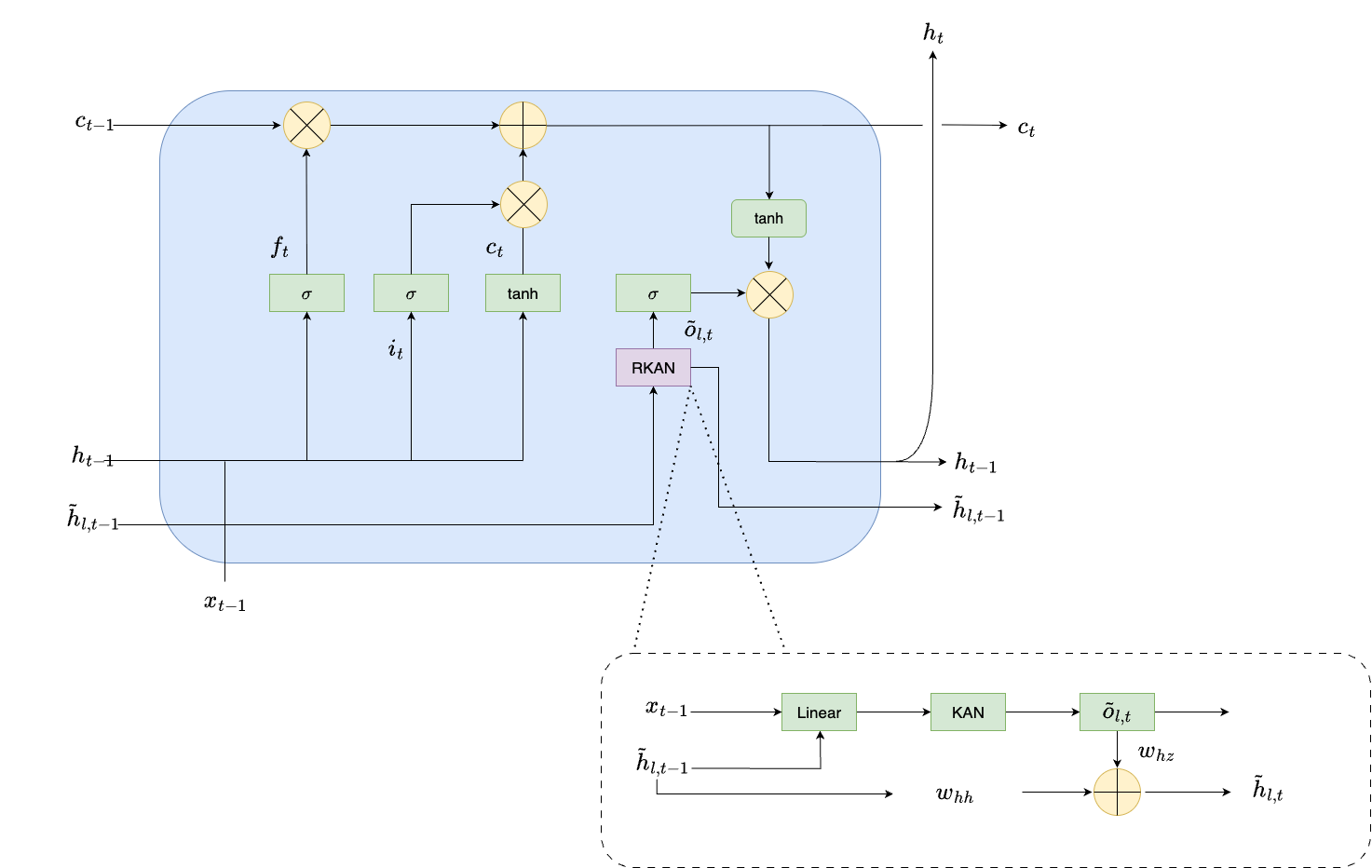}
    \caption{Temporal Kolmogorov-Arnold Networks (TKAN)}
    \label{fig:temporal_kan}
\end{figure}

\end{abstract}

\section{Introduction}

Time series forecasting is an important branch of statistical analysis and machine learning. The development of machine learning models has accelerated rapidly in the past few years. Time series, which can be defined as sequences of data indexed in time, are essential in finance, meteorology, and even in the field of healthcare. The ability to accurately predict the future evolution of such data has become a strategic issue for many different industries that are constantly seeking innovation. In recent years, the growing interest in this area of research is mainly due to the massive increase in the availability of data. Moreover, the increased computer processing capacity now makes it possible to process large datasets. As a second driver for innovation, new statistical methods, deep learning techniques \cite{sezer2020financial} and hybrid models have offered new opportunities to improve the accuracy and efficiency of forecasts. Besides econometric models such as ARMA/ARIMA (\cite{makridakis1997arma,mondal2014study}) and their numerous extensions, Recurrent Neural Networks (RNNs) \cite{medsker2001recurrent} yield families of models that have been recognized for their proven effectiveness in terms of forecasting \cite{hewamalage2021recurrent}.

\medskip

RNNs have been proposed to address the "persistence problem", i.e. the potential dependencies between the successive observations of some time series. Therefore, RNNs most often outperform "static" networks as MLPs ~\cite{le2015simple}. Traditional methods of gradient descent may not be sufficiently effective for training Recurrent Neural Networks (RNNs), particularly in capturing long-term dependencies ~\cite{learning_lt}. Meanwhile, ~\cite{grn_evaluation} conducted an empirical study revealing the effectiveness of gated mechanisms in enhancing the learning capabilities of RNNs. Actually, RNNs have proved to be one of the most powerful tools for processing sequential data and solving a wide range of difficult problems in the fields of automatic natural language processing, translation, image processing and time series analysis where MLPs \cite{hornik1989multilayer,haykin1998neural,cybenko1989approximation} cannot perform well. When it comes to sequential data management, MLPs face limitations. Unlike RNNs, MLPs, are not designed to manage sequential data, information only flows in one direction, from input to output (feedforward connection). This specifica tion makes them not suitable for modeling temporal sequences where taking into account sequential patterns is essential for prediction. Another weakness of MLPs is that these networks have no embedded mechanism for cell memory state management. Recurrent Neural Networks (RNNs) \cite{werbos1990backpropagation,williams1989learning} have provided an answer to these problems. However, standard RNNs have solved one problem but created another: "vanishing" or "exploding gradient problem" \cite{hochreiter1998vanishing}. Traditional methods of gradient descent may not be sufficiently effective for training RNNs, particularly in capturing long-term dependencies \cite{learning_lt}. Meanwhile, \cite{grn_evaluation} conducted an empirical study revealing the effectiveness of gated mechanisms in enhancing the learning capabilities of RNNs. Actually, RNNs have proved to be one of the most powerful tools for processing sequential data and solving a wide range of difficult problems in the fields of automatic natural language processing, translation, and time series analysis. Recently, Liu et al. \cite{liu2024kan} proposed Kolmogorov-Arnold Networks (KANs) as an alternative to MLPs. The Kolmogorov-Arnold Network (KAN) is an efficient new neural network architecture known for its improved performance and interpretability. Unlike traditional models, KANs apply activation functions on the connections between nodes, and these functions can even learn and adapt during training. In addition to using KAN Layer, \textit{Temporal Kolmogorov-Arnold Networks (TKANs)} manage temporality within a data sequence. The idea is to design a new family of neural networks capable of catching long-term dependency. Our primary idea is to introduce an external memory module which can be attached to KAN Layers. This "memory" can store information that is relevant to the temporal context and can be accessed by the network during processing. This allows the network to explicitly learn and utilize past information. Codes are available at \href{https://github.com/remigenet/TKAN}{TKAN repository} and can be installed using the following command: \codeword{pip install tkan}. Data are accessible if the reader wishes to reproduce our experiments using the GitHub link provided above.

\section{Kolmogorov-Arnold Networks (KANs)}

Multi-Layer Perceptrons (MLPs) \cite{hornik1989multilayer} are extension of original perceptron proposed by \cite{rosenblatt1958perceptron}. They were inspired by the universal approximation theorem \cite{cybenko1989approximation} which states that a feed-forward network (FFN) with a single hidden layer containing a finite number of neurons can arbitrarily well approximate any continuous functions on a compact subset of $\Rb^n$. On the other side, Kolmogorov-Arnold Networks (KAN) focuses on the Kolmogorov-Arnold representation theorem \cite{kolmogorov1961representation}. The Kolmogorov-Arnold representation theorem states that any multivariate continuous function $f$ can be represented as a composition of univariate functions and through additive operations: 
\begin{equation}
    f(x_1, \dots, x_n) = \sum_{q=1}^{2n+1} \Phi_q \left( \sum_{p=1}^n \phi_{q,p}(x_p) \right)
    \label{eq:KART_tkan}
\end{equation}
where $\phi_{q,p}$ are univariates functions that map each input variable $x_p \in [0,1]$ to $ \Rb$, and $\Phi_q : \Rb \rightarrow \Rb$. Since all functions to be learnt are univariate, we can parametrize every 1D function as a B-spline curve. The learnable coefficients are then associated with local B-spline basis functions. The key insight comes when we see the similarities between MLPs and KAN. In MLPs, a layer includes a linear transformation followed by nonlinear operations, and you can make the network deeper by adding more layers. A KAN layer is rather
\begin{align}
    {\mathbf\Phi}=\{\phi_{q,p}\},\quad p=1,2,\ldots,n_{\rm in},\quad q=1,2\ldots,n_{\rm out},
\end{align}
where $\phi_{q,p}$ are parametrized function of learnable parameters. In the Kolmogorov-Arnold theorem, the inner functions form a KAN layer with $n_{\rm in}=n$ and $n_{\rm out}=2n+1$, and the outer functions form a KAN layer with $n_{\rm in}=2n+1$ and $n_{\rm out}=1$. So, the Kolmogorov-Arnold representations in eq.~(\ref{eq:KART_tkan}) are simply compositions of two KAN layers. Now it becomes clear what it means to have Deep Kolmogorov-Arnold representation. Taking the notation from \cite{liu2024kan} let us define a shape of KAN $[n_0,n_1,\cdots,n_L],$ where $n_i$ is the number of nodes in the $i^{\rm th}$ layer of the computational graph. We denote the $i^{\rm th}$ neuron in the $l^{\rm th}$ layer by $(l,i)$, and the activation value of the $(l,i)$-neuron by $x_{l,i}$. Between layer $l$ and layer $l+1$, there are $n_ln_{l+1}$ activation functions: the activation function that connects $(l,i)$ and $(l+1,j)$ is denoted by 
\begin{align}
    \phi_{l,j,i},\quad l=0,\cdots, L-1,\quad i=1,\cdots,n_{l},\quad j=1,\cdots,n_{l+1}.
\end{align}
The pre-activation of $\phi_{l,j,i}$ is simply $x_{l,i}$; the post-activation of $\phi_{l,j,i}$ is denoted by $\tilde{x}_{l,j,i}\equiv \phi_{l,j,i}(x_{l,i})$. The activation value of the $(l+1,j)$ neuron is simply the sum of all incoming post-activations: 
\begin{equation}\label{eq:kanforward}
    x_{l+1,j} =  \sum_{i=1}^{n_l} \tilde{x}_{l,j,i} = \sum_{i=1}^{n_l}\phi_{l,j,i}(x_{l,i}), \qquad j=1,\cdots,n_{l+1}.
\end{equation}
Rewriting it under the matrix form will give:
\begin{equation}\label{eq:kanforwardmatrix}
    \mathbf{x}_{l+1} = 
    \underbrace{\begin{pmatrix}
        \phi_{l,1,1}(\cdot) & \phi_{l,1,2}(\cdot) & \cdots & \phi_{l,1,n_{l}}(\cdot) \\
        \phi_{l,2,1}(\cdot) & \phi_{l,2,2}(\cdot) & \cdots & \phi_{l,2,n_{l}}(\cdot) \\
        \vdots & \vdots & & \vdots \\
        \phi_{l,n_{l+1},1}(\cdot) & \phi_{l,n_{l+1},2}(\cdot) & \cdots & \phi_{l,n_{l+1},n_{l}}(\cdot) \\
    \end{pmatrix}}_{\mathbf{\Phi}_l}
    \mathbf{x}_{l},
\end{equation}
where ${\mathbf \Phi}_l$ is the function matrix corresponding to the $l^{\rm th}$ KAN layer. A general KAN network is a composition of $L$ layers: given an input vector  $\mathbf{x}_0\in\mathbb{R}^{n_0}$, the output of KAN is:
\begin{equation}\label{eq:KAN_forward}
    {\rm KAN}(\mathbf{x}) = (\mathbf{\Phi}_{L-1}\circ \mathbf{\Phi}_{L-2}\circ\cdots\circ\mathbf{\Phi}_{1}\circ\mathbf{\Phi}_{0})\mathbf{x}.
\end{equation}

\section{Temporal Kolmogorov-Arnold Networks (TKANs)}

\begin{figure}[H]
    \centering
    \includegraphics[width=0.8\linewidth]{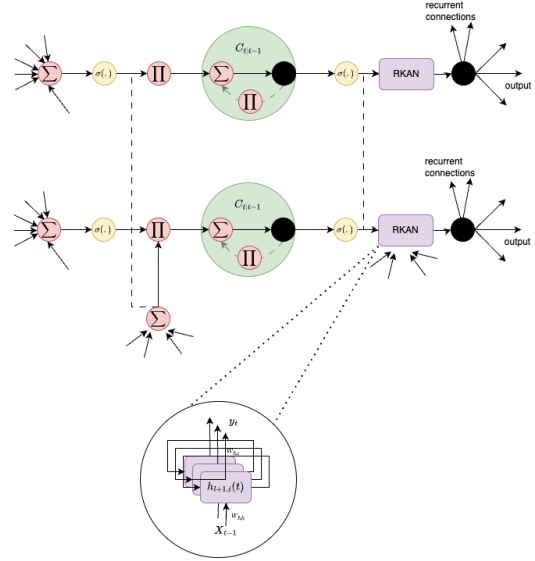}
    \caption{Temporal Kolmogorov-Arnold Networks (TKAN)}
    \label{fig:temporal_kan}
\end{figure}

After proposing the RKAN (Recurrent Kolmogorov-Arnold Network), which integrates temporality management by adapting the concept of Kolmogorov-Arnold networks to temporal sequences, we developed an additional innovation to build our neural network: the TKAN layer. This TKAN layer combines the RKAN architecture with a slightly modified LSTM (Long Short-Term Memory) cell. The idea is to propose an extension of the model proposed by \cite{liu2024kan} to manage sequential data and temporality during the learning task. The objective is to provide a framework for time series forecasting on multiple steps ahead. As discussed previously, RNNs' weakness generally lies in the difficulty of capturing the persistence of information when some input sequence is quite long, resulting in a significant loss of information in some tasks. LSTMs address this problem through the use of a gating mechanism. LSTMs can be computationally more expensive than standard RNNs; however, this extra complexity is often justified by better performance on complex learning tasks. The integration of an LSTM cell combined with the RKAN enables the capture of complex nonlinearities with learnable activation functions of RKAN, but also the maintenance of a memory of past events over long periods with the LSTM cell architecture. This combination offers superior modeling power for tasks involving complex sequential data. The major components of the TKAN are:

\begin{itemize}
    \item \textbf{RKAN Layers:} RKAN layers enable the retention of short term memory from previous states within the network. Each RKAN layer manages this short term memory throughout the processing in each layer.
    \item \textbf{Gating Mechanisms:} these mechanisms help to manage the information flow. The model decides which information should be retained or forgotten over time.
\end{itemize}

\subsection{Recurring Kolmogorov-Arnold Networks (RKAN)}

In neural networks, particularly in the context of recurrent neural networks (RNN), a recurrent kernel refers to the set of weights that are applied to the hidden state from the previous timestep during the network’s operation. This kernel plays a crucial role in how an RNN processes sequential data over time. Let us denote $\tau=1,2,...$ some discrete time steps. Each step has a forward pass and backward pass. During the forward pass, the output or activation of units are computed. During the backward pass, the computation of the error for all weights is made. During each timestep, an RNN receives an input vector and the hidden state from the previous timestep $h_{t-1}$ . The recurrent kernel is a matrix of weights that transforms this previous hidden state. This operation is usually followed by an addition; the result of this transformation is passed through a non-linear function, an activation function $f(.)$, that can take many forms: tanh, ReLU etc.
The update stage could be formulated as: 
\begin{equation}
\label{eq:rnn_update}
    h_t = f(W_{hh} h_{t-1} + W_{hx} x_t + {b_h}),
\end{equation}
where $h_t$ is the hidden state at time $t \in \tau$; $W_h$ and $W_x$ are the recurrent kernel, a weight matrix that transforms the previous hidden states $h_{t-1}$, and the input kernel, a weight matrix transforming the current input denoted $x_t$, respectively. In the next sections, we propose a new way of updating KANs. We propose a process to maintain the memory of past inputs by incorporating previous hidden states into the current states, enabling the network to exhibit dynamic temporal behavior. Recurrent Kernel is the key so that RKAN layers learn from sequences for which context and order matter. We design the TKAN to leverage the power of Kolmogorov-Arnold Network while offering memory management to handle time dependency. To introduce time dependency, we modify each transformation function $\phi_{l}$. Let's denote the sublayers' memory state $\tilde{h}_{l,t}$, initialized with zeros and of shape $(\text{KAN}_{out})$.
The input that is fed to each sub KAN layers in the RKAN are created by doing:
\begin{equation}
    s_{l,t}=W_{l,\tilde{x}} x_t + W_{l,\tilde{h}} \tilde{h}_{l,t-1},
\end{equation}
where $W_{l,\tilde{x}}$ is the weight of the $l$-th layer applied to $x_t$ which is the input at time $t$. Moreover, $W_{l,\tilde{h}}$ are the weights of the $l$-th layer applied to its previous substate. Let us first denote $\text{KAN}_{in}$ and $\text{KAN}_{out}$, the input dimension of RKAN Layer input and outputs, respectively.
The shape of $W_{l,\tilde{x}} \in \Rb^{(d, \text{KAN}_{in})} $ and $W_{l,\tilde{h}} \in \Rb^{(\text{KAN}_{out}, \text{KAN}_{in})}$, which leads to $s_{l,t}^{\text{KAN}_{in}}$. Set
\begin{equation}
\tilde{o}_{t} =  \phi_{l}(s_{l,t}),
\end{equation}
where $\phi_l$ is a KAN layer. The "memory" step \(\tilde{h}_{l,t}\) is defined as a combination of past hidden states, such as:
\begin{equation}
\tilde{h}_{l,t} =  W_{hh} \tilde{h}_{l,t-1} + W_{hz} \tilde{o}_{t},
\label{eq:update_state_tkan}
\end{equation}
introducing vectors of weights, that weight the importance of past values relative to the most recent inputs. 

\subsection{TKAN Architecture}

\begin{figure}[H]
    \centering
    \includegraphics[width=0.8\linewidth]{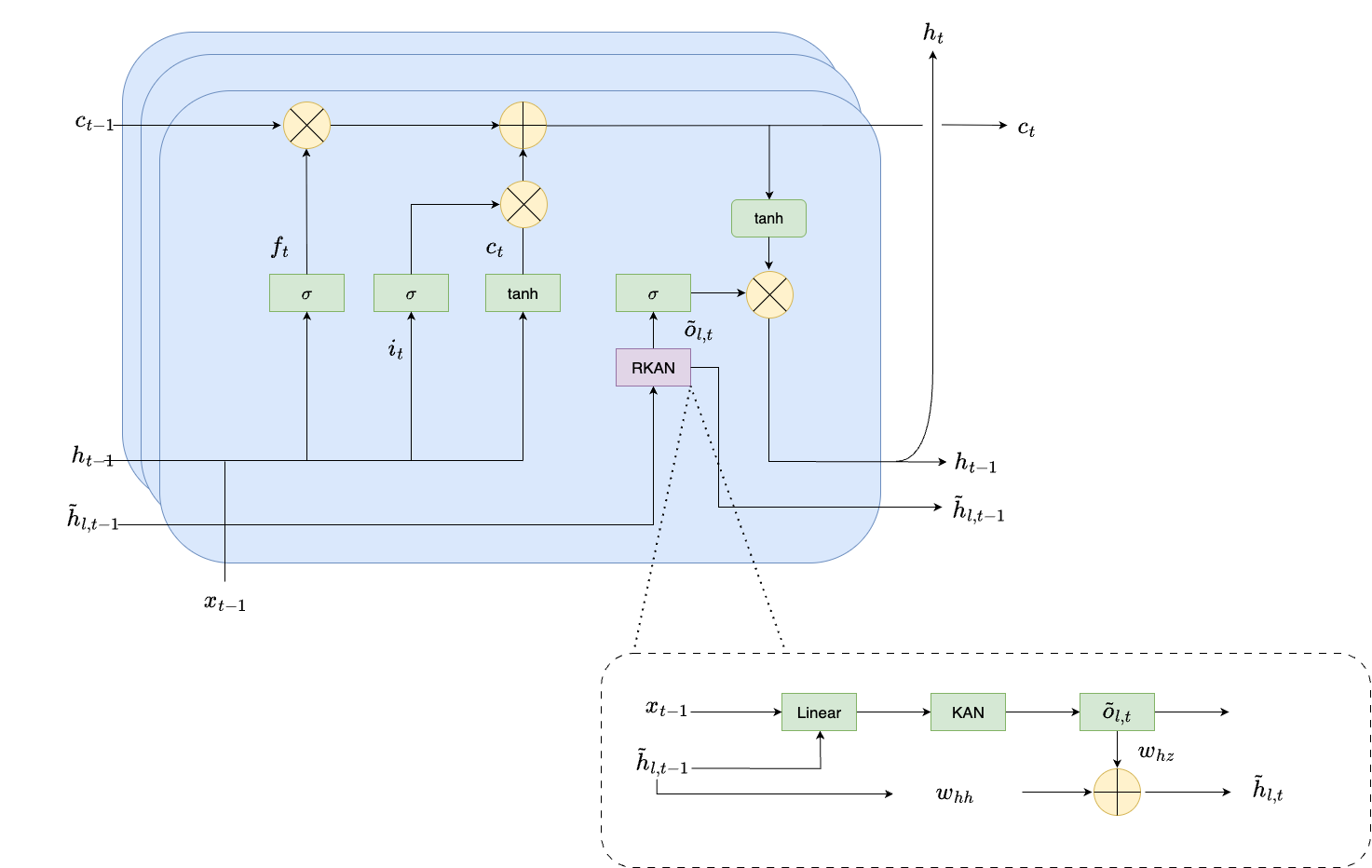}
    \caption{A three layers Temporal Kolmogorov-Arnold Networks (TKAN) Block}
    \label{fig:tkan}
\end{figure}

For the next step, our aim is to manage memory and we took inspiration from LSTM \cite{hochreiter1997long,staudemeyer2019understanding}.  We denote $x_t$ the input vector of dimension $d$. This unit uses several internal vectors and gates to manage information flow. The forget gate with activation vector $f_t$ given by
\begin{equation}
    f_t = \sigma(W_f x_t + U_f h_{t-1} + b_f),
\end{equation}
decides what information to forget from the previous state. 
The input gate, with activation vector denoted $i_t$, with 
\begin{equation}
    i_t = \sigma(W_i x_t + U_i h_{t-1} + b_i),
\end{equation}
controls which new information to include. 
The output gate, with activation vector $o_t$. Set
\begin{equation}
    r_t = \text{Concat}[\phi_1(s_{1,t}),\phi_2(s_{2,t}),...,\phi_L(s_{L,t})],
    \label{eq:concat_kan_out_tkan}
\end{equation}
where $r_t$ is a concatenation of the output of multiple KAN Layers, and
\begin{equation}
    o_t = \sigma(W_{o}r_t + b_o),
    \label{eq:out_tkan}
\end{equation}
where $W_{o} \in \Rb^{(\text{KAN}_{out} * L,out)}$, $out$ denoting the output dimension of TKAN.     
Equation \eqref{eq:out_tkan} determines what information from the current state to output given $r_t$ given from \eqref{eq:concat_kan_out_tkan}. $\tilde{h}_t$ is the "sub" memory of the RKAN layers. The hidden state of the TKAN layer, $h_t$, captures the unit's output, while the cell state $c_t$ is updated as follows:
\begin{equation}
    c_t = f_t \odot c_{t-1} + i_t \odot \tilde{c}_t,
\end{equation}
where $\tilde{c}_t = \sigma(W_c x_t + U_c h_{t-1} + b_c)$
represents its internal memory. All these internal states have a dimensionality of $h$.
The output of the final hidden layer, denoted $h_t$, is given by:
\begin{equation}
    h_t = o_t \odot \tanh(c_t).
    \label{eq:hidden_update}
\end{equation}
In the following section, we will describe the learning task and proceed with several tests. We started to test the power of prediction of our model for one step ahead forecasting and in the second time for multi-step ahead \cite{fan2019multi,lim2021temporal}.

\section{Learning task}
In order to assess the relevance of extending our model, we created a simple training task to judge its ability to improve the out-of-sample prediction accuracy over several steps ahead for standard layers such as GRU or LSTM.
We have chosen to carry out our study not on synthetic data, but rather on real market data, as this seems more relevant to us. Indeed, synthetic market data can be easily biased to match an experiment.

\subsection{Task Definition and Dataset}

The task is to predict the market notional trades over future periods. This is recognized as a difficult task as the market behavior is difficult to predict. Nonetheless, this should not be impossible as, unlike returns, volumes have internal patterns such as seasonality, autocorrelation, and so on. We used \href{www.binance.com}{Binance} as our only data source, as it is the largest player in the crypto-currency market. Moreover, due to the lack of an overall regulation that would take into account all exchange data, exchanges are subject to a lot of falsified data by small players who use wash trading to boost their figures.

\medskip

Our dataset consists of the notional amounts traded each hour on several assets:  BTC, ETH, ADA, XMR, EOS, MATIC, TRX, FTM, BNB, XLM, ENJ, CHZ, BUSD, ATOM, LINK, ETC, XRP, BCH and LTC, which are to be used to predict just one of them, BTC. The data period runs from January 1, 2020 to December 31, 2022. 

\subsection{Preprocessing}
Data preparation is a necessary step for most machine learning methods, in order to help gradient descent calculations when data are of different sizes, to obtain stationarity series, etc.
This is even more true when using the Kolmogorov-Arnold network, as the underlying B-Spline activation functions exhibit power exponent. Thus, having poorly scaled data would result in over or underflows that would hinder learning. This is also true for market volume data, or notional data. Indeed, they are series with very different scales across assets and between points in time, not to mention non-stationarity issues over a long period.

\medskip

To obtain data that can be used for training, but also return meaningful losses to optimize, we use a two-stage scaling.
The first step is to divide the values in the series by the moving median of the last two weeks. This moving median window is also shifted by the number of steps forward we want to predict, so as not to include foresight. This first pre-treatment aims to make the series more stationary over time.
The second pre-processing we apply is a simple MinMaxScaling per asset: even if the minimum of the series is 0, this way of working is simply a matter of dividing by its maximum value. The objective is to scale the data in the $[0, 1]$ interval to avoid an explosive effect during learning due to the power exponent. This pre-processing is, however, adjusted on the training set, the adjustment meaning only the search for the maximum value of each series, which is then used directly on the test set. This means that, on the test set, it is possible to obtain observations that are greater than 1. As long as no optimization is launched, this is not a problem. Finally, we split our dataset into a training set and a test set, with a standard proportion of 80-20. This represents over 21,000 points in the training set and 5,000 in the test set. 

\subsection{Loss Function for Model Training}
Since we have a numerical prediction problem, we have opted to optimize our model using the root mean square error (RMSE) as the loss function, whose formula is simple: RMSE is the square root of
$$
\text{MSE} = \frac{1}{N} \sum_{i=1}^{N} \left(\hat{X}_{t+1}^{(i)} - X_{t+1}^{(i)}\right)^2,
$$
where \(N\) represents the number of samples in the dataset, \(\hat{X}_{t+1}^{(i)}\) denotes the predicted notional values of Bitcoin at time \(t+1\) for the \(i\)-th sample, and the quantities \(X_{t+1}^{(i)}\) are the corresponding true values. We invoke RMSE first because it is the most widely used and standard loss in machine learning for this type of problem.
The second reason is related to the metric we want to use to display the results, namely the R-squared (\(R^2\)). Indeed, (\(R^2\)) is interesting as a metric because it not only gives information on the error but also on the error given the variance of the estimated series, which means it's much easier to tell whether the model is performing well or not. This explains why it is a measure widely used by econometricians and practitioners. 
However, minimizing MSE is exactly the same problem as maximizing the (\(R^2\)), as its formula indicates:
\[
R^2 = 1 - \frac{\sum_{i=1}^{N} (\hat{X}_{t+1}^{(i)} - X_{t+1}^{(i)})^2}{\sum_{i=1}^{N} (X_{t+1}^{(i)} - \bar{X}_{t+1})^2},
\]

\subsection{Benchmarks}

\subsubsection{Model Architectures}
In order to compare with comparable things, we tested our TKAN layers against two of the most widely used RNNs for multi-step predictions, namely gated recurrent units (GRU) and long short-term memory (LSTM). As can be seen, we are not comparing ourselves to complete model architectures such as a temporal fusion transformer, as what we are proposing is more a layer than a complete model architecture. 

\medskip

To fairly compare the three models, we have opted for a very simple configuration. We create three models to be compared in the same way:
\begin{enumerate}
	\item An initial recurrent layer of 100 units that returns complete sequences,
	\item An intermediate recurrent layer of 100 units, which returns only the last hidden state,
	\item A final dense layer with linear activation, with as many units as there are timesteps to predict ahead
\end{enumerate}
For the TKAN model, we used 5 B-spline activations of order 0 to 4 as sublayer activations, while we used the standard activation function for GRU and LSTM.
Finally, we also compared the three models to the most naive benchmark, which consists of using the last value as a predictor of the future, the value being repeated when using several predictions in advance.

\subsubsection{Note on training details}

Metrics are calculated directly on scaled data and not on unscaled data. There are two reasons for this: firstly, MinMax scaling has no impact on the metric since the minimum is 0 and the data interval is $[0,1]$; rescaling would not have changed the R-squared.
Rescaling from the median split would have caused the series mean to be unstable over time. This would have resulted in the error magnitude drifting for certain parts of the series, making the metric unreliable or meaningless. In terms of optimizing model details, we used the Adam optimizer, one of the most popular choices, used 20\% of our training set as a validation set and included two training recalls. The first is an early learning stopper, which interrupts training after 6 consecutive periods without improvement on the validation set, and restores the weights associated with the best loss obtained on the validation set. The second is a plateau learning rate reduction, which halves the learning rate after three consecutive periods showing no improvement on the validation set.

\subsection{Results}

To evaluate the model's performance, taking into account the risk of poor adaptation that may occur, we repeat the experiment 5 times for each, and display below the mean and standard deviation of the train results obtained from these 5 experiments.

\subsubsection{Performance Metrics Summary}

\begin{table}[H]
\centering
\caption{Average (\(R^2\)) obtained over 5 runs}
\begin{tabular}{ccccc}
\toprule
Time & TKAN:5 B-Spline & GRU:default & LSTM:default & Last Value \\
\midrule
1  & 0.33736 & 0.365136 & 0.355532 & 0.292171 \\
3  & 0.21227 & 0.200674 & 0.061220 & -0.062813 \\
6  & 0.13784 & 0.082504 & -0.225838 & -0.331346 \\
9  & 0.09803 & 0.087164 & -0.290584 & -0.457718 \\
12 & 0.10401 & 0.017864 & -0.473220 & -0.518252 \\
15 & 0.09512 & 0.033423 & -0.404432 & -0.555633 \\
\bottomrule
\end{tabular}
\end{table}

\begin{table}[H]
    \centering
    \caption{Standard Deviation of the (\(R^2\)) obtained over 5 runs}
    \begin{tabular}{ccccc}
    \toprule
    Time & TKAN:5 B-Spline & GRU:default & LSTM:default \\
    \midrule
    1  & 0.00704 & 0.008336 & 0.011163 \\
    3  & 0.00446 & 0.004848 & 0.080200 \\
    6  & 0.01249 & 0.023637 & 0.062710 \\
    9  & 0.02430 & 0.014833 & 0.052729 \\
    12 & 0.00132 & 0.086386 & 0.085746 \\
    15 & 0.00701 & 0.024078 & 0.092729 \\
    \bottomrule
    \end{tabular}
\end{table}

The results show a very logical decrease in terms of R2 with the number of forward steps ahead, which is quite normal since we have less information for the forward steps.
However, the results clearly show two things: the first is that while all models have a relatively small difference in performance over a short horizon such as 1 to 3 steps ahead, the relative performances change much more with a longer time horizon.
The LSTM model even becomes useless and counter-productive with 6 periods, while TKAN and GRU always achieve a higher average R-squared value. However, TKAN stands out with longer time steps, with an R-squared value at least 25\% higher than that of GRU. Another very interesting point is model stability, i.e. the ability to calibrate well-functioning weights from samples without too much variation from one experiment to the next, and here again, TKAN showed much better stability than all the other models.

\subsubsection{Training Dynamics and Model Stability}
To better understand the differences in performance between the models, we visualized training and validation losses over several training sessions for each model. These graphs offer a dynamic view of each model's learning process and ability to generalize beyond the training data.

\begin{figure}[H]
  \centering 
  \includegraphics[width=1\columnwidth]{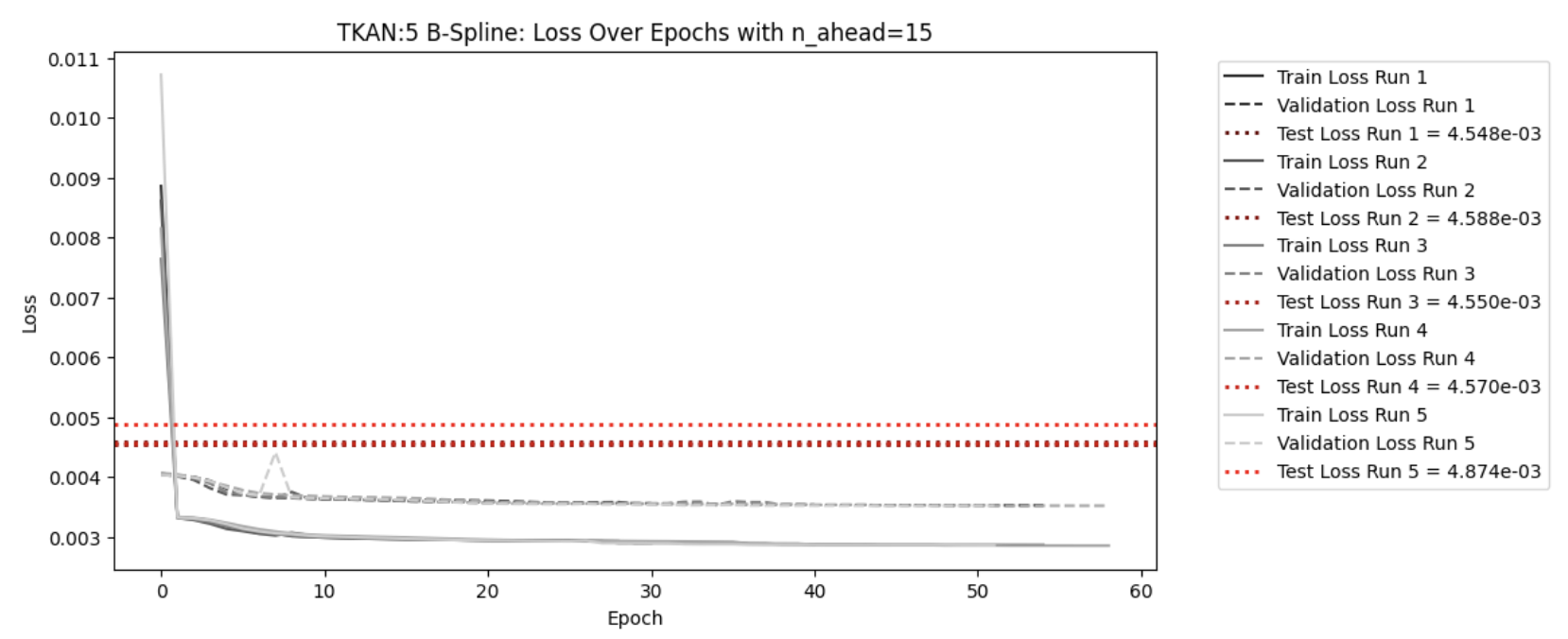}
  \caption{TKAN training and validation loss over epochs}
  \label{fig:tkan_loss}
\end{figure}

\begin{figure}[H]
  \centering 
  \includegraphics[width=1\columnwidth]{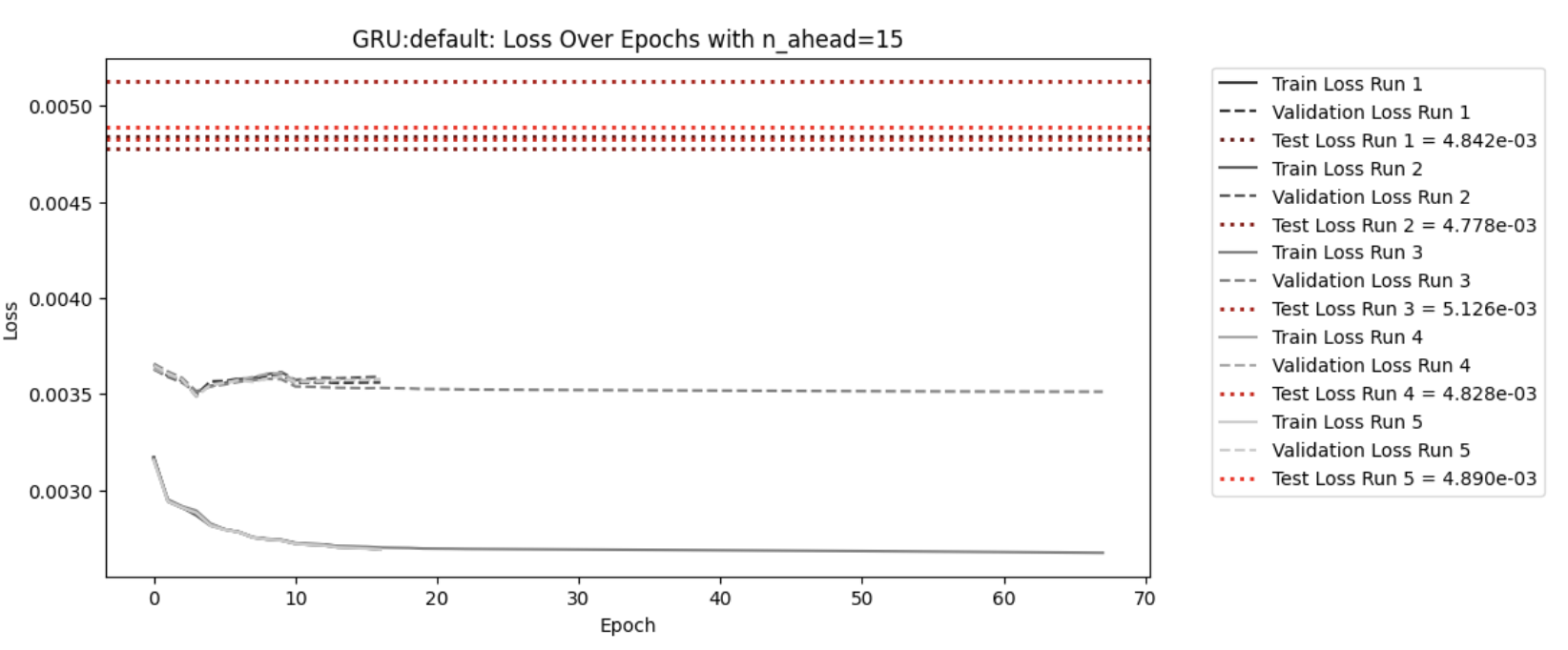}
  \caption{GRU training and validation loss over epochs}
  \label{fig:gru_loss}
\end{figure}

\begin{figure}[H]
  \centering 
  \includegraphics[width=1\columnwidth]{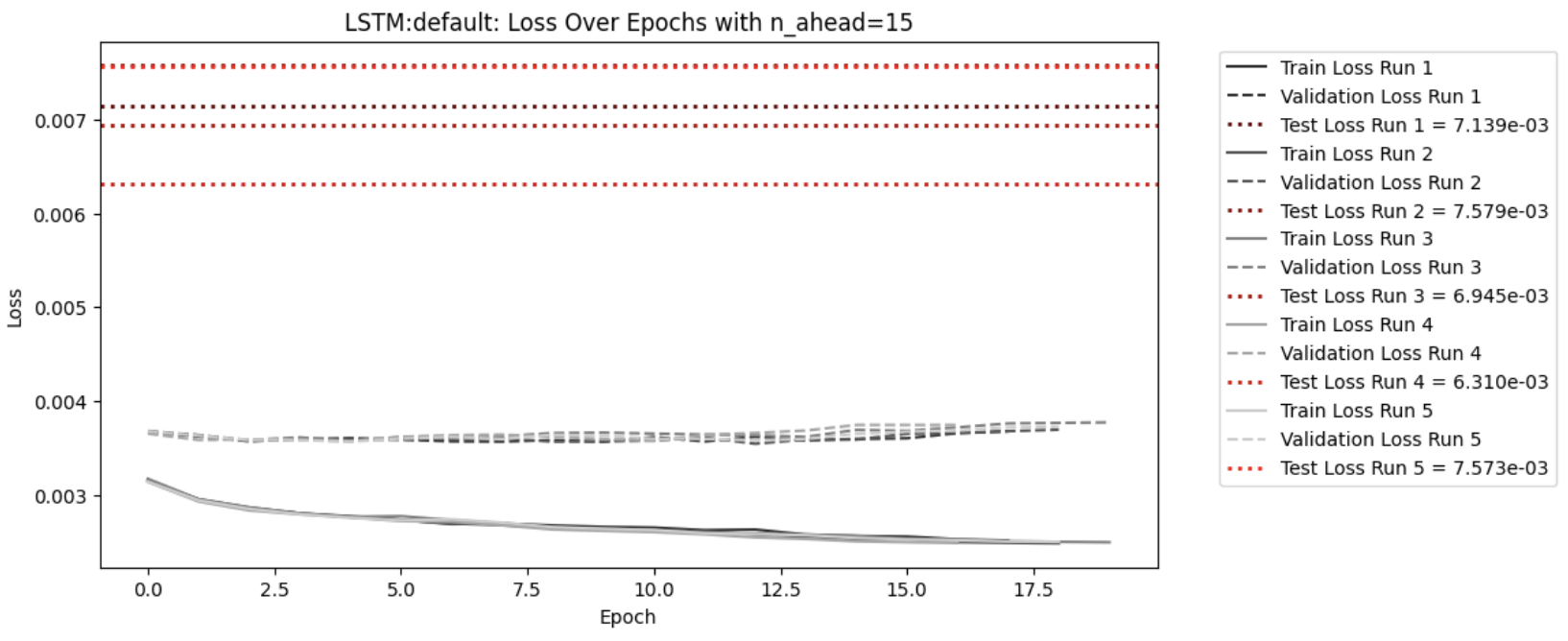}
  \caption{LSTM training and validation loss over epochs}
  \label{fig:lstm_loss}
\end{figure}

The visual representations clearly corroborate the statistical results presented above. The GRU and LSTM models show a significant divergence between their learning loss and validation trajectories, particularly as the number of epochs increases. This divergence suggests a potential overfitting where the model learns idiosyncrasies from the training data rather than generalizing them. This stability in the TKAN model's learning process, evident in the closer alignment of its learning and validation loss curves, implies a consistent learning model that effectively captures the underlying patterns in the data without overfitting. 

\section{Conclusion}
In this paper, we proposed an adaptation of the Kolmogorov-Arnold Network architecture for time series that incorporates both recurring and gating mechanisms. The architecture, while not so complicated, enables improving multiple steps' performances and stability compared to traditional methods and seems to be promising. The temporal Kolmogorov-Arnold networks (TKANs) combine the best features of recurrent neural networks (RNNs) and Kolmogorov-Arnold Networks (KANs). This new architecture tackles the usual problems of RNNs (long-term dependency). TKANs embed Recurrent Kolmogorov-Arnold Networks (RKAN). These layers help the system to memorize and use new and old information efficiently. Compared to more traditional models such as LSTM and GRU, TKAN particularly stands out when it comes to making longer-term predictions, showing that it is capable of handling different situations and longer periods of time. Our experiments show that it is usable and more stable than GRU and LSTM on real historical market data. While not specifically interesting for short-term predictions, it especially demonstrates an ability to largely outperform other models when it comes to multi-step predictions.This also confirms that the idea developed in the original KAN paper works well on real use cases and is totally relevant for time series analysis. This paper opens interesting new ways to improve our capacities to calibrate accurate time-series models over multiple steps, which is one of the hardest tasks in temporal analysis.

\bibliographystyle{IEEEtran}
\bibliography{bib}

\end{document}